\documentclass[preprint,12pt]{elsarticle}

\usepackage{amssymb}
\usepackage{amsmath}

\usepackage{booktabs}
\usepackage{multirow}
\usepackage{graphicx}
\usepackage{makecell}
\usepackage{placeins}
\usepackage{float}
\usepackage[ruled,vlined,linesnumbered]{algorithm2e}
\SetKwInput{KwInput}{Input}
\SetKwInput{KwOutput}{Output}
\SetNlSty{textbf}{}{:\ }
\SetNlSkip{0.8em}

\journal{Information Processing & Management}

\begin{document}

\begin{frontmatter}

\title{SketchGraphNet: A Memory-Efficient Hybrid Graph Transformer for Large-Scale Sketch Corpora Recognition}

\author[inst1]{Shilong Chen} 
\ead{shilong_chen0109@foxmail.com}

\author[inst2]{Mingyuan Li}
\ead{mingyuan1007_li@foxmail.com}

\author[inst1]{Zhaoyang Wang}
\ead{z.y.wang@stu.qhnu.edu.cn}

\author[inst1]{Zhonglin Ye\corref{cor1}}
\ead{zhonglin_ye@foxmail.com}

\author[inst3]{Haixing Zhao}
\ead{h.x.zhao@163.com}

\cortext[cor1]{Corresponding author}

\affiliation[inst1]{organization={College of Computer, Qinghai Normal University},
            city={Xining},
            postcode={810001}, 
            state={Qinghai},
            country={China}}

\affiliation[inst2]{organization={College of Computer, Nanjing University}, 
            city={Nanjing},
            postcode={210023}, 
            state={Jiangsu},
            country={China}}

\affiliation[inst3]{organization={College of Computer, Qinghai Minzu University}, 
            city={Xining},
            postcode={810007}, 
            state={Qinghai},
            country={China}}

\begin{abstract}
This work investigates large-scale sketch recognition from a graph-native perspective, where free-hand sketches are directly modeled as structured graphs rather than raster images or stroke sequences. We propose SketchGraphNet, a hybrid graph neural architecture that integrates local message passing with a memory-efficient global attention mechanism, without relying on auxiliary positional or structural encodings. To support systematic evaluation, we construct SketchGraph, a large-scale benchmark comprising 3.44 million graph-structured sketches across 344 categories, with two variants (A and R) to reflect different noise conditions. Each sketch is represented as a spatiotemporal graph with normalized stroke-order attributes. On SketchGraph-A and SketchGraph-R, SketchGraphNet achieves Top-1 accuracies of 83.62\% and 87.61\%, respectively, under a unified training configuration. MemEffAttn further reduces peak GPU memory by over 40\% and training time by more than 30\% compared with Performer-based global attention, while maintaining comparable accuracy.
\end{abstract}


\end{frontmatter}



\section{Introduction}
\label{sec1}
Graph neural networks (GNNs) have become a standard tool for learning from structured relational data and have shown strong performance across a range of graph learning applications~\cite{li2025kolmogorov, wozniak2025accurate, zhang2025graph, li2025graph, liu2025multignn}. However, most message-passing architectures rely on localized neighborhood aggregation, which limits their ability to capture long-range dependencies when information propagation is constrained by graph connectivity~\cite{togninalli2019wasserstein}.

Free-hand sketches encode human concepts through sparse strokes and temporal drawing processes. While most existing sketch recognition approaches rely on rasterized images or sequential stroke modeling, these representations discard explicit structural information inherent in the drawing process. From an information processing perspective, sketches can be naturally interpreted as structured graph objects, where nodes correspond to sampled stroke points and edges encode local geometric continuity.

This work adopts a graph-native modeling paradigm, in which sketches are directly represented and processed as graphs without intermediate image rendering or sequence-only abstraction. Under this formulation, sketch recognition becomes a structured graph learning problem, raising challenges in large-scale training efficiency, global dependency modeling, and numerical robustness.

To mitigate the receptive field bottleneck of message-passing GNNs, the Graph Transformer paradigm integrates Transformer-style self-attention into graph modeling, facilitating global information exchange. Representative architectures such as Graphormer~\cite{ying2021transformers}, GraphGPS~\cite{rampavsek2022recipe}, and Exphormer~\cite{shirzad2023exphormer} enhance long-range interactions through explicit positional or structural encodings (PE/SE). However, these designs often impose substantial computational and memory overhead, and their performance is sensitive to the choice and calibration of encoding schemes.

Recent studies, including SGFormer~\cite{wu2023sgformer}, AnchorGT~\cite{zhu2024anchorgt}, and FlashLinearAttention~\cite{beck2025tiled}, have investigated more efficient attention formulations or simplified global modeling strategies. Achieving a robust balance between accuracy, efficiency, and numerical stability remains challenging in large-scale settings~\cite{errica2019fair}. Furthermore, standard graph benchmarks like ZINC and MUTAG are limited in scale and evaluation consistency, failing to adequately reflect the complexities of realistic large-scale deployment scenarios~\cite{bechler2025position}.

Free-hand sketches represent a unique class of sparse graph data. While they offer an intuitive representation of human concepts, sketches exhibit distinct structural characteristics: extreme topological sparsity, non-deterministic stroke ordering, and substantial noise. These properties pose severe challenges for conventional CNN- and RNN-based approaches, which struggle to capture the underlying structural dependencies of sketch data~\cite{ha2017neural}. Despite growing interest in sketch understanding, systematic graph-based modeling of sketches, together with large-scale and unified evaluation frameworks, remains noticeably absent, which impedes rigorous assessment of structured learning methods in this domain.

From an engineering perspective, scaling Transformer-based architectures to large-scale settings is often hindered by numerical instabilities, particularly during attention computation under mixed-precision training~\cite{team2025kimi}. In hybrid models combining global self-attention with local graph convolution, such instabilities commonly manifest as Inf or NaN values caused by uncontrolled Query--Key interactions. Prior work has attempted to mitigate this issue through explicit stabilization strategies, including logit soft-capping, QK-Norm, and QK-Clip~\cite{team2024gemma, team2025gemma, team2025kimi}. However, these approaches typically introduce additional constraints or tuning complexity. This underscores the necessity for attention mechanisms that are inherently more numerically stable while remaining compatible with resource-constrained large-scale training.

In this work, the term large-scale denotes the number of training samples rather than the size of individual graphs. Each sketch graph is constructed with a fixed number of uniformly sampled points, shifting the primary scalability challenges to training throughput, memory efficiency, and numerical robustness. To address these challenges, we exploit the inherent temporal order of sketching as an inductive bias, capturing sequential drawing logic without relying on auxiliary positional encoding schemes.

In contrast to prior approaches, this work adopts an orthogonal stabilization strategy at the feature-transformation and implementation level, combining memory-efficient self-attention with intrinsic temporal cues of sketch data. This design improves numerical robustness and scalability without introducing additional encoding complexity.

\subsection*{Research Objectives}

The objective of this study is to investigate whether large-scale free-hand sketch recognition can be effectively addressed from a purely graph-native perspective. Specifically, we aim to:

(i) establish a unified, large-scale benchmark for graph-structured sketches that enables controlled evaluation under corpus-scale training settings;

(ii) design a hybrid local--global graph architecture that preserves structural modeling advantages while remaining memory-efficient and numerically stable under mixed-precision training;

(iii) empirically assess whether graph-native modeling can achieve competitive accuracy--efficiency trade-offs compared with raster-based, sequential, and Transformer-style baselines under identical experimental configurations.

These objectives frame the methodological and experimental design choices throughout the paper.

In summary, sketch-graph learning faces three key challenges: 
(i) the lack of a unified, large-scale benchmark for graph-structured sketches; 
(ii) memory and efficiency constraints that hinder the scalability of Graph Transformer models; and 
(iii) the absence of lightweight architectures that jointly support local--global modeling and numerical stability. 
To address these challenges, we introduce \textbf{SketchGraphNet}, a hybrid graph neural network that integrates local message passing with a global attention mechanism tailored for large-scale sketch classification. The main contributions of this work are as follows:

\begin{itemize}
    \item \textbf{Large-scale, graph-native sketch benchmark.} We construct \textbf{SketchGraph}, a graph-structured sketch benchmark comprising 344 categories and 3.44 million samples. Each node is associated with spatial coordinates and a temporal attribute, and two dataset variants are provided to evaluate robustness under different noise conditions.
    
    \item \textbf{Memory-efficient attention design.} We propose \textbf{MemEffAttn}, a numerically stable and memory-efficient global attention module that significantly reduces computational overhead when integrated into hybrid GNN architectures.
    
    \item \textbf{Lightweight local--global fusion without PE/SE.} By exploiting intrinsic temporal information as an inductive bias, SketchGraphNet achieves effective local--global interaction without relying on auxiliary structural or positional encoding modules.
    
    \item \textbf{Empirical validation at scale.} Extensive experiments demonstrate that SketchGraphNet achieves consistently higher classification accuracy and improved efficiency compared with representative baselines on large-scale sketch datasets.
\end{itemize}

By jointly addressing benchmark availability, architectural efficiency, and numerical robustness, SketchGraphNet provides a scalable and practical framework for large-scale graph-based sketch understanding.

\begin{figure}[H]
\centering
\includegraphics[width=\linewidth]{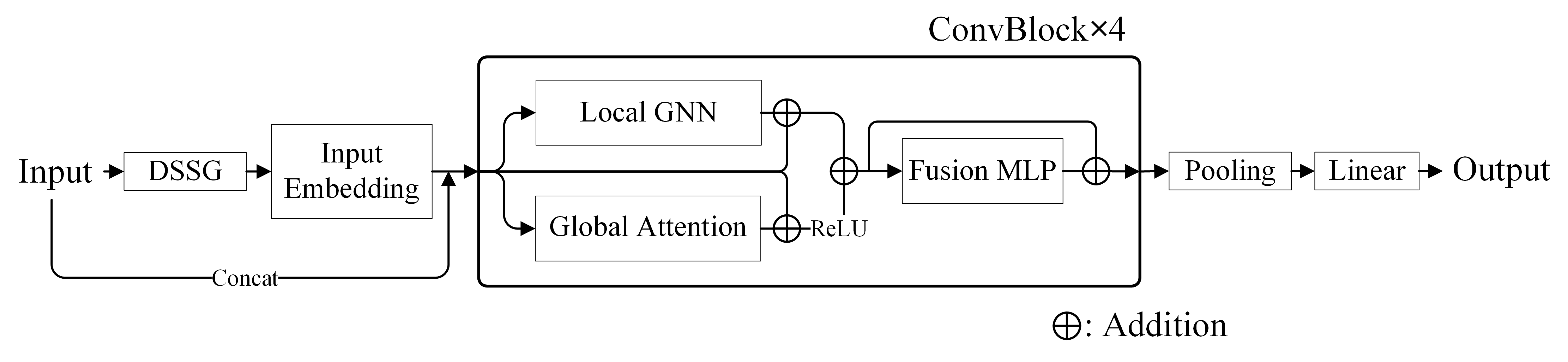}
\caption{Architecture of SketchGraphNet.}\label{fig:1}
\end{figure}

\section{Methods}
\label{sec2}

\subsection{Notation and Overall Architecture}
\label{subsec:notation_overview}

Let a sketch graph be denoted as $G=(V,E)$, with $n=|V|$ nodes and $m=|E|$ edges. Each node is associated with an input feature vector, and the node feature matrix is written as $\boldsymbol{X}\in\mathbb{R}^{n\times d_{\mathrm{in}}}$, where $d_{\mathrm{in}}$ denotes the input feature dimension. The model uses a hidden dimension $d$, and the node representation at layer $l$ is denoted by $\boldsymbol{H}^{(l)}\in\mathbb{R}^{n\times d}$.

\textbf{SketchGraphNet} is built upon the GraphGPS framework and adopts a hybrid architecture that integrates local message passing with global self-attention. The model consists of a GIN-based local GNN branch and a memory-efficient global attention module, enabling joint modeling of local stroke topology and global interactions in sketch graphs. Compared with the original GraphGPS design, SketchGraphNet introduces efficiency-oriented refinements to better accommodate the spatiotemporal characteristics of free-hand sketches. Figure~\ref{fig:1} provides an overview of the architecture, and Algorithm~\ref{alg:sketchgraphnet} summarizes the end-to-end forward propagation.

\begin{algorithm}[H]
\caption{SketchGraphNet Forward Propagation}
\label{alg:sketchgraphnet}
\DecMargin{1em}
\KwInput{
Sketch graph $G=(V,E)$ with node features $\mathbf{X}\in\mathbb{R}^{n\times d_{\mathrm{in}}}$; 
number of ConvBlocks $L$; 
model parameters $\Theta$
}
\KwOutput{
Predicted class probabilities $\hat{\mathbf{y}}$
}

\IncMargin{1em}

\BlankLine
\If{DSSG is enabled}{
    $E \leftarrow \mathrm{DSSG}(G)$ \tcp*{edge structure enhancement}
}

$\mathbf{H}^{(0)} \leftarrow \mathrm{ChebConv}(\mathbf{X};\Theta_{\mathrm{cheb}})$ \tcp*{input embedding}

\For{$l=1$ \KwTo $L$}{
    $\mathbf{H}^{(l)}_{\mathrm{loc}} \leftarrow \mathrm{LocalGNN}(\mathbf{H}^{(l-1)}; \Theta^{(l)}_{\mathrm{loc}})$\;
    
    $\mathbf{H}^{(l)}_{\mathrm{glob}} \leftarrow \mathrm{MemEffAttn}(\mathbf{H}^{(l-1)}; \Theta^{(l)}_{\mathrm{att}})$\;
    
    $\mathbf{Z}^{(l)} \leftarrow 
    \sigma\!\left(\mathbf{H}^{(l)}_{\mathrm{glob}} + \mathbf{H}^{(l-1)}\right)
    + \left(\mathbf{H}^{(l)}_{\mathrm{loc}} + \mathbf{H}^{(l-1)}\right)$\;
    
    $\mathbf{H}^{(l)} \leftarrow 
    \mathrm{BatchNorm}\!\left(\mathbf{W}^{(l)}_f \mathbf{Z}^{(l)} + \mathbf{b}^{(l)}_f\right)
    + \mathbf{Z}^{(l)}$\;
}

$\mathbf{g} \leftarrow \mathrm{GlobalPooling}(\mathbf{H}^{(L)})$\;

$\hat{\mathbf{y}} \leftarrow \mathrm{Softmax}(\mathbf{W}_{\mathrm{cls}}\mathbf{g} + \mathbf{b}_{\mathrm{cls}})$\;

\Return $\hat{\mathbf{y}}$
\end{algorithm}

\subsection{Local--Global Hybrid Convolution Block}
\label{subsec:hybrid_block}

As shown in Fig.~\ref{fig:1}, DSSG~\cite{chen2024dssg} is used as an edge-structure enhancement module for sketch graphs. After this refinement, the initial node representations are obtained by projecting the input features into a latent space using a Chebyshev convolution~\cite{defferrard2016convolutional}:
\begin{equation}
\boldsymbol{H}^{(0)}=\mathrm{ChebConv}\!\left(\boldsymbol{X};\boldsymbol{\Theta}_{\mathrm{cheb}}\right)\in\mathbb{R}^{n\times d},
\label{eq:1}
\end{equation}
where $\boldsymbol{\Theta}_{\mathrm{cheb}}$ denotes the learnable parameters of the Chebyshev filter.

The backbone of \textbf{SketchGraphNet} consists of $L$ stacked convolutional blocks. Each block jointly models local and global interactions through a local GNN operator and a global attention module. The local branch adopts GINConv~\cite{xu2018powerful} with a two-layer MLP to update node representations. For the $l$-th block ($l=1,\dots,L$), the local and global outputs are computed as
\begin{equation}
\begin{aligned}
\boldsymbol{H}_{\mathrm{loc}}^{(l)} &=
\mathrm{GINConv}\!\left(
\boldsymbol{H}^{(l-1)};\,
\boldsymbol{\Theta}_{\mathrm{gin}}^{(l)}
\right),\\
\boldsymbol{H}_{\mathrm{glob}}^{(l)} &=
\mathrm{MemEffAttn}\!\left(
\boldsymbol{H}^{(l-1)};\,
\boldsymbol{\Theta}_{\mathrm{att}}^{(l)}
\right),
\end{aligned}
\label{eq:2}
\end{equation}
where $\boldsymbol{\Theta}_{\mathrm{gin}}^{(l)}$ and $\boldsymbol{\Theta}_{\mathrm{att}}^{(l)}$ denote the learnable parameters of the local message-passing and global attention modules.

The outputs of the local and global branches are fused through a gated residual formulation:
\begin{equation}
\boldsymbol{Z}^{(l)}=
\sigma\!\left(
\boldsymbol{H}_{\mathrm{glob}}^{(l)}+\boldsymbol{H}^{(l-1)}
\right)
+
\left(
\boldsymbol{H}_{\mathrm{loc}}^{(l)}+\boldsymbol{H}^{(l-1)}
\right),
\label{eq:3}
\end{equation}
where $\sigma(\cdot)$ denotes the ReLU function. The fused features are then normalized and projected with a residual connection:
\begin{equation}
\boldsymbol{H}^{(l)}=
\mathrm{BatchNorm}\!\left(
\boldsymbol{W}_{f}^{(l)}\boldsymbol{Z}^{(l)}+\boldsymbol{b}_{f}^{(l)}
\right)
+
\boldsymbol{Z}^{(l)}.
\label{eq:4}
\end{equation}

After the final convolutional block, node representations are aggregated by a global pooling operator to produce a graph-level embedding, which is then passed to a linear classifier for sketch category prediction.

Compared with the standard GraphGPS framework, which combines local and global representations by simple concatenation or summation, SketchGraphNet employs an explicit nonlinear gating formulation to integrate global attention outputs with residual information from previous layers. This formulation operates at the feature fusion level and is orthogonal to logit-level stabilization techniques such as QK-Norm~\cite{team2025gemma} and QK-Clip~\cite{team2025kimi}.

\begin{figure}[t]
\centering
\includegraphics[width=\linewidth]{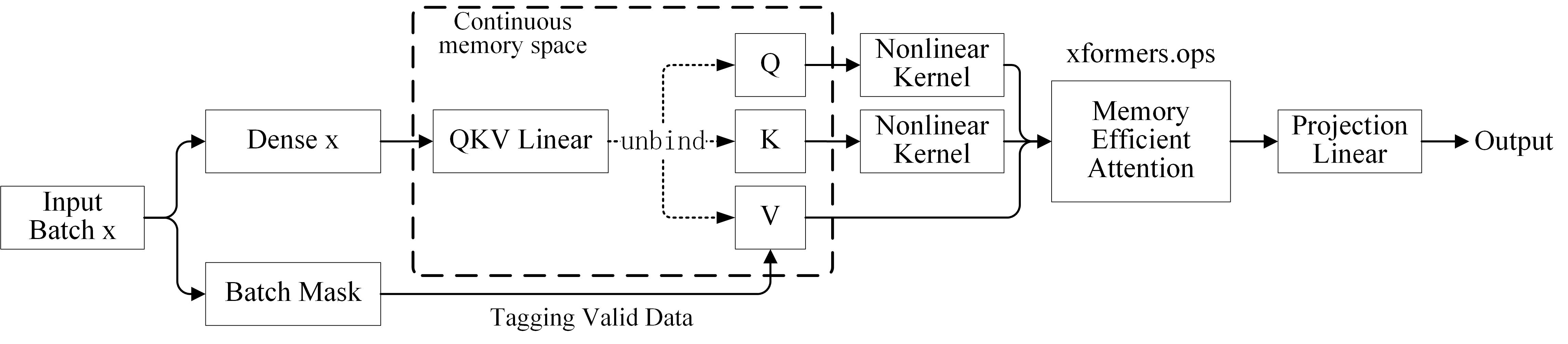}
\caption{Architecture of MemEffAttn.}\label{fig:2}
\end{figure}

\begin{algorithm}[H]
\caption{Memory-Efficient Global Attention (MemEffAttn)}
\label{alg:memeffattn}
\KwInput{
Dense node features $\mathbf{H}\in\mathbb{R}^{B\times N\times d}$; 
batch mask $\mathbf{M}\in\{0,1\}^{B\times N}$
}
\KwOutput{
Updated node features $\mathbf{H}'\in\mathbb{R}^{B\times N\times d}$
}

\BlankLine
$\mathbf{Q},\mathbf{K},\mathbf{V} \leftarrow \mathrm{Linear}_{QKV}(\mathbf{H})$ \tcp*{QKV projection}

$\mathbf{Q} \leftarrow \phi(\mathbf{Q}),\;
 \mathbf{K} \leftarrow \phi(\mathbf{K})$ \tcp*{non-negative kernel (ReLU)}

$\mathbf{A} \leftarrow 
\mathrm{MemoryEfficientAttention}(\mathbf{Q},\mathbf{K},\mathbf{V},\mathbf{M})$ \tcp*{xFormers tiled Softmax}

$\mathbf{H}' \leftarrow \mathrm{Linear}_{\mathrm{proj}}(\mathbf{A})$\;

\Return $\mathbf{H}'$
\end{algorithm}

\subsection{Memory-Efficient Global Attention}
\label{subsec:memeffattn}

\subsection{Computational Considerations}
\label{subsec:complexity}

The proposed MemEffAttn module is summarized in Algorithm~\ref{alg:memeffattn}. It applies a non-negative mapping to queries and keys and computes exact Softmax attention using tiled execution provided by xFormers.

\subsubsection{Standard Self-Attention}
\label{subsec:standard_attn}

In graph learning tasks, the computational and memory cost of standard self-attention scales quadratically with the number of nodes. Given node representations $\boldsymbol{H}\in\mathbb{R}^{n\times d}$, standard self-attention is defined as
\begin{equation}
\mathrm{Attn}(\boldsymbol{H})=
\operatorname{softmax}\!\left(
\frac{\boldsymbol{Q}\boldsymbol{K}^{\top}}{\sqrt{d_h}}
\right)\boldsymbol{V},
\label{eq:5}
\end{equation}
where $\boldsymbol{Q}=\boldsymbol{H}\boldsymbol{W}_Q$, $\boldsymbol{K}=\boldsymbol{H}\boldsymbol{W}_K$, and $\boldsymbol{V}=\boldsymbol{H}\boldsymbol{W}_V$ denote the query, key, and value projections, and $d_h$ is the per-head feature dimension.

\subsubsection{MemEffAttn with Non-negative Feature Mapping}
\label{subsec:kernel_mapping}

For the global attention branch, SketchGraphNet replaces the Performer layer in the original GraphGPS architecture with the proposed \textbf{MemEffAttn} module. MemEffAttn computes Softmax attention exactly, while forming attention logits using a non-negative feature mapping applied to the query and key projections:
\begin{equation}
\mathrm{MemEffAttn}(\boldsymbol{H})=
\operatorname{softmax}\!\left(
\frac{\phi(\boldsymbol{Q})\,\phi(\boldsymbol{K})^{\top}}{\sqrt{d_h}}
\right)\boldsymbol{V},
\label{eq:6}
\end{equation}

where $\phi(\cdot)$ denotes an element-wise non-negative mapping implemented by the ReLU function. The Softmax operation itself is computed exactly; the modification is limited to the formation of attention logits through $\phi(\boldsymbol{Q})$ and $\phi(\boldsymbol{K})$.

This formulation is distinct from kernelized or linear attention, as Softmax normalization is preserved and no low-rank or random-feature approximation is introduced.

The non-negativity of $\phi(\cdot)$ reshapes the distribution of query--key interactions and improves numerical robustness under mixed-precision training. Unlike stabilization techniques that operate directly on logits or optimization dynamics (e.g., explicit clipping or QK-level constraints), this design applies a feature-space transformation prior to attention computation and does not introduce additional optimizer-side modifications.

SketchGraphNet does not rely on auxiliary positional or structural encoding modules; global dependencies are inferred directly through self-attention over node features.

\subsection{Implementation Details}
\label{subsec:impl}

\subsubsection{Dense Batching and Masking}
\label{subsec:dense_batching}

As shown in Fig.~\ref{fig:2}, irregular graph inputs are converted into dense tensor representations by aligning multiple graphs in a batch to fixed-size feature matrices. To handle varying graph sizes, each batch is padded to the maximum number of nodes $n_{\mathrm{max}}$ in the batch, and graphs with fewer nodes are filled with zero-valued entries. A binary mask is used to mark padded positions, ensuring that these entries are excluded from attention computation.

Within the aligned representation, the mapping $\phi(\cdot)$ is applied element-wise to the query ($\boldsymbol{Q}$) and key ($\boldsymbol{K}$) projections before attention computation.

During training, cross-entropy loss with label smoothing is used as the optimization objective and is applied consistently across all compared models.

\subsubsection{Blockwise Attention via xFormers}
\label{subsec:xformers_blockwise}

From an implementation perspective, \textbf{MemEffAttn} is implemented using the tiled, blockwise attention interface provided by the xFormers library~\cite{lefaudeux2022xformers}. This implementation computes Softmax attention in blocks without explicitly materializing the full $n\times n$ attention matrix, thereby reducing peak memory usage during training. Unlike approximation-based efficient attention methods such as Performer~\cite{choromanski2020rethinking} and Cosformer~\cite{qin2022cosformer}, MemEffAttn does not introduce random feature projections or frequency-based transformations, but relies on implementation-level memory optimization. Figure~\ref{fig:2} provides a schematic overview of the MemEffAttn module.

Given the scale of the sketch-graph datasets considered in this work, computational and memory efficiency are critical for practical model scalability. In standard self-attention, computing attention weights incurs a time complexity of $\mathcal{O}(n^2 d)$ and requires $\mathcal{O}(n^2)$ memory to store the attention matrix, where $n$ denotes the number of nodes and $d$ the hidden dimension. This quadratic dependence on $n$ restricts the feasible batch size and limits the scalability of graph Transformer models in large-scale settings.

SketchGraphNet adopts a tiled, blockwise attention computation strategy. Although the theoretical time complexity remains $\mathcal{O}(n^2 d)$, blockwise execution avoids explicit materialization of the full $n\times n$ attention matrix. As a result, peak memory usage during attention computation is bounded by the size of local blocks rather than the full matrix.

In our setting, sketch graphs are standardized to a fixed number of nodes ($n=100$) through uniform point sampling. Under this fixed-size representation, blockwise execution results in a stable and bounded memory footprint across batches. Quantitative evaluations of training latency and memory usage are reported in Section~\ref{subsec4.4} (Table~\ref{tab:2}), where the proposed attention module and its implementation are compared with representative baselines.

\section{Dataset Construction}
\label{sec3}

\subsection*{Dataset Overview}

SketchGraph contains 3.44 million graph-structured sketches across 344 semantic categories, with 10,000 samples per category. Each sketch is converted into a spatiotemporal graph representation with fixed node count ($n=100$) obtained via uniform point sampling along strokes. Two variants are provided:

\begin{itemize}
    \item \textbf{Version A:} unfiltered sketches directly derived from QuickDraw;
    \item \textbf{Version R:} sketches verified as recognizable by the QuickDraw recognition system.
\end{itemize}

All experiments reported in this paper are conducted exclusively on SketchGraph under identical data splits and preprocessing configurations.

\subsection*{Source Data and Category Processing}

We construct \textbf{SketchGraph}, a large-scale benchmark composed of graph-structured free-hand sketches. The dataset supports systematic evaluation of graph-based sketch understanding methods under high-throughput training settings. Graph construction follows the pipeline introduced in DSSG~\cite{chen2024dssg}, ensuring consistent and reproducible conversion from raw sketch inputs to graph representations.

SketchGraph is derived from the QuickDraw dataset~\cite{ha2017neural}. The ``line'' category is excluded due to its trivial topology and limited stroke complexity. The remaining 344 categories are processed using balanced sampling to reduce class imbalance and ensure uniform category coverage.

\subsection*{Graph Construction Protocol}

For each sketch, a graph representation is constructed following the DSSG procedure, producing node and edge structures that preserve the spatial and temporal characteristics of the drawing process. In total, SketchGraph contains 3.44 million graph-structured sketch samples. All samples are stored using binary serialization with offset-based indexing. This format supports parallel random access during data loading and high-throughput I/O in large-scale training.

Each node in \textbf{SketchGraph} is associated with an explicit temporal attribute $t$ that encodes stroke order. The temporal value is derived from normalized timestamps. Given the timestamp sequence $T=(t_1,\dots,t_n)$ of the $n$ sampled nodes in a sketch, we compute $t_{\max}=\max_i t_i$ and normalize each timestamp as $t_i' = t_i / t_{\max}$. Each node is thus represented by a three-dimensional feature vector $\boldsymbol{x}_i=[x_i, y_i, t_i']$, where $(x_i,y_i)$ denotes the absolute coordinates on a $256\times256$ pixel canvas.

In \textbf{SketchGraph}, each sketch is represented as a collection of disconnected subgraphs, with each subgraph corresponding to a single stroke in QuickDraw. For each stroke, a fixed number of points is uniformly sampled along its trajectory, and the sampled points are connected in temporal order to form a simple path graph without self-loops. Under this construction, edges are defined only between adjacent sampled points within the same stroke, preserving the geometric continuity of the drawing. This representation converts sequential point samples into a graph format without introducing additional topological assumptions beyond the stroke structure.

\clearpage
\begin{figure}[p]
\centering
\includegraphics[width=\linewidth,
                height=0.95\textheight,
                keepaspectratio]{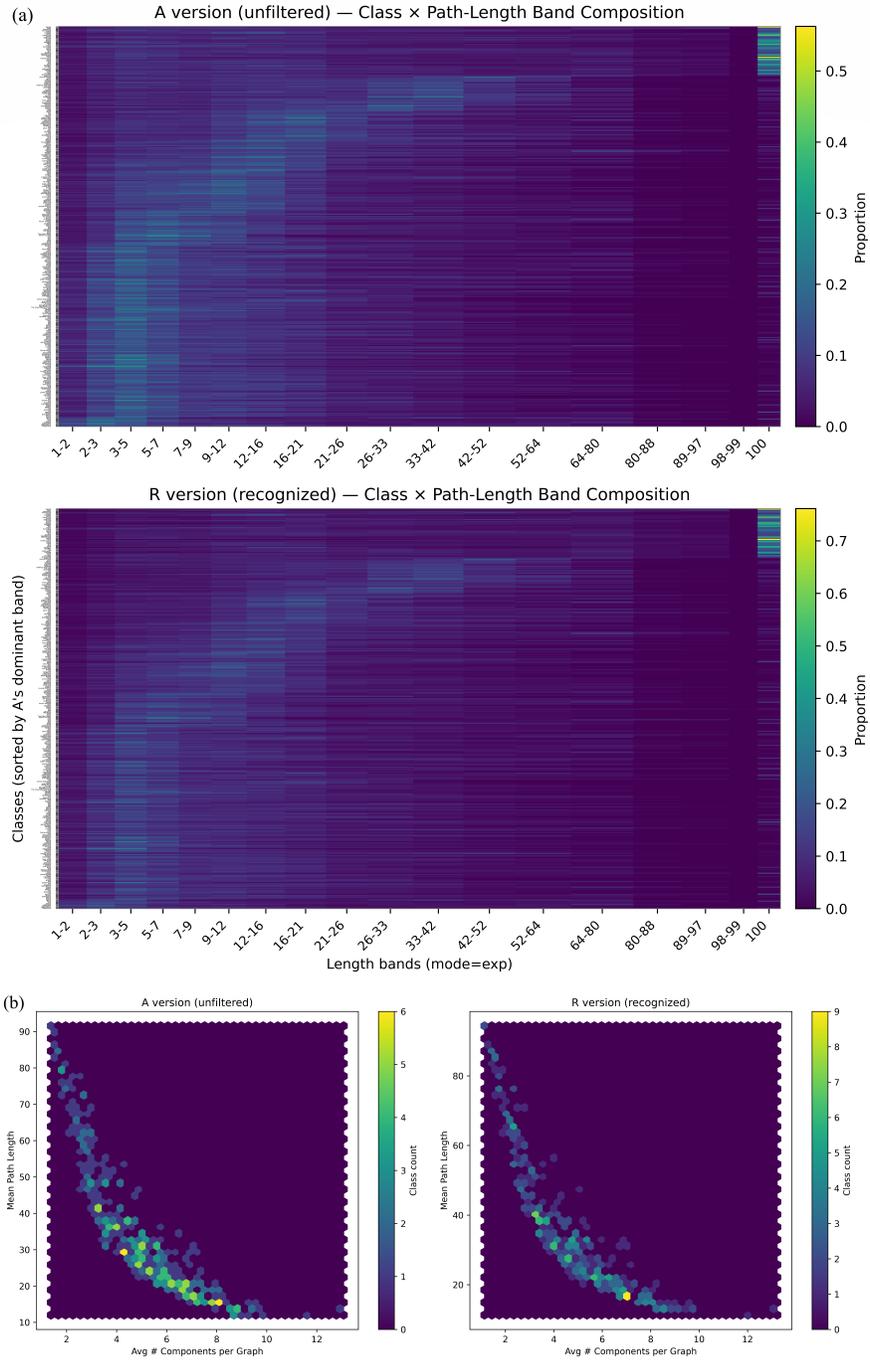}
\caption{Statistical Comparison Between the A and R Versions of the SketchGraph Dataset: (a) Path-Length Distributions Across Categories; (b) Category-Level Density in the Space of Stroke Count Versus Path Length.}
\label{fig:3}
\end{figure}
\clearpage

\subsection*{Dataset Variants and Structural Statistics}

The dataset is released in two variants to support evaluation under different noise conditions. The \textbf{A} version contains unfiltered sketches, while the \textbf{R} version includes only sketches verified as recognizable by the built-in QuickDraw recognition system. Both variants share the same category set and graph construction protocol, with 10{,}000 samples per class.

Figure~\ref{fig:3} summarizes the structural differences between the two dataset variants. As shown in Fig.~\ref{fig:3}(a), the R version exhibits a more concentrated path-length distribution, with fewer short or fragmented paths. Figure~\ref{fig:3}(b) shows the category-level distribution in the space of average stroke count and average path length, where the R version occupies a more concentrated region.

\begin{figure}[htbp]
\centering
\includegraphics[width=\linewidth]{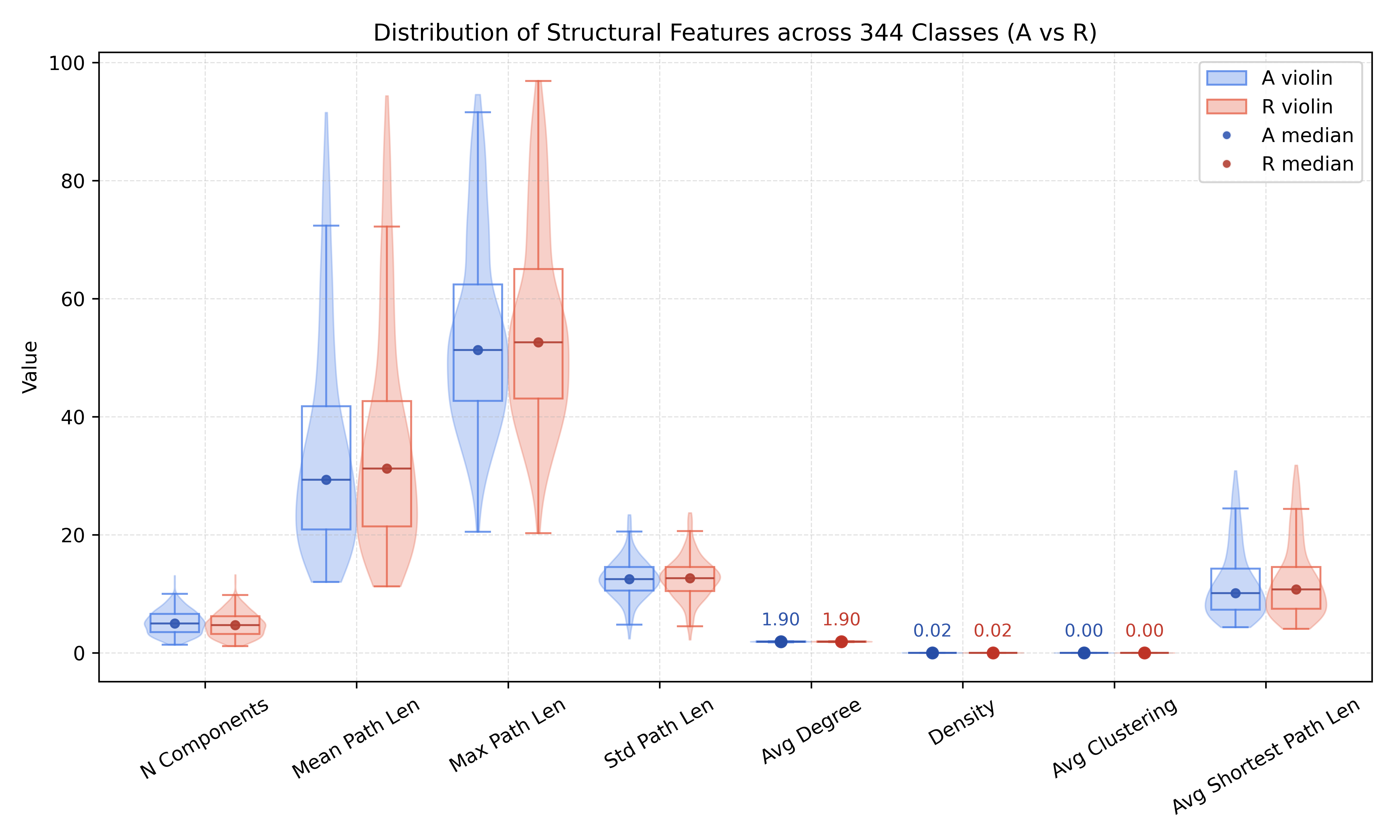}
\caption{Structural Statistics Aggregated Across All Categories in the Dataset.}\label{fig:4}
\end{figure}

To further characterize the structural properties of \textbf{SketchGraph}, Fig.~\ref{fig:4} reports several graph-level statistics aggregated over all 344 categories. The distributions show clear differences between Version A and Version R in terms of structural concentration and variability.

These statistics describe the sparsity patterns and connectivity characteristics of large-scale sketch graphs. Differences between the two dataset variants are observable across multiple graph-level measures, providing a structural reference for future graph-based sketch modeling.

\begin{figure}[htbp]
\centering
\includegraphics[width=\linewidth]{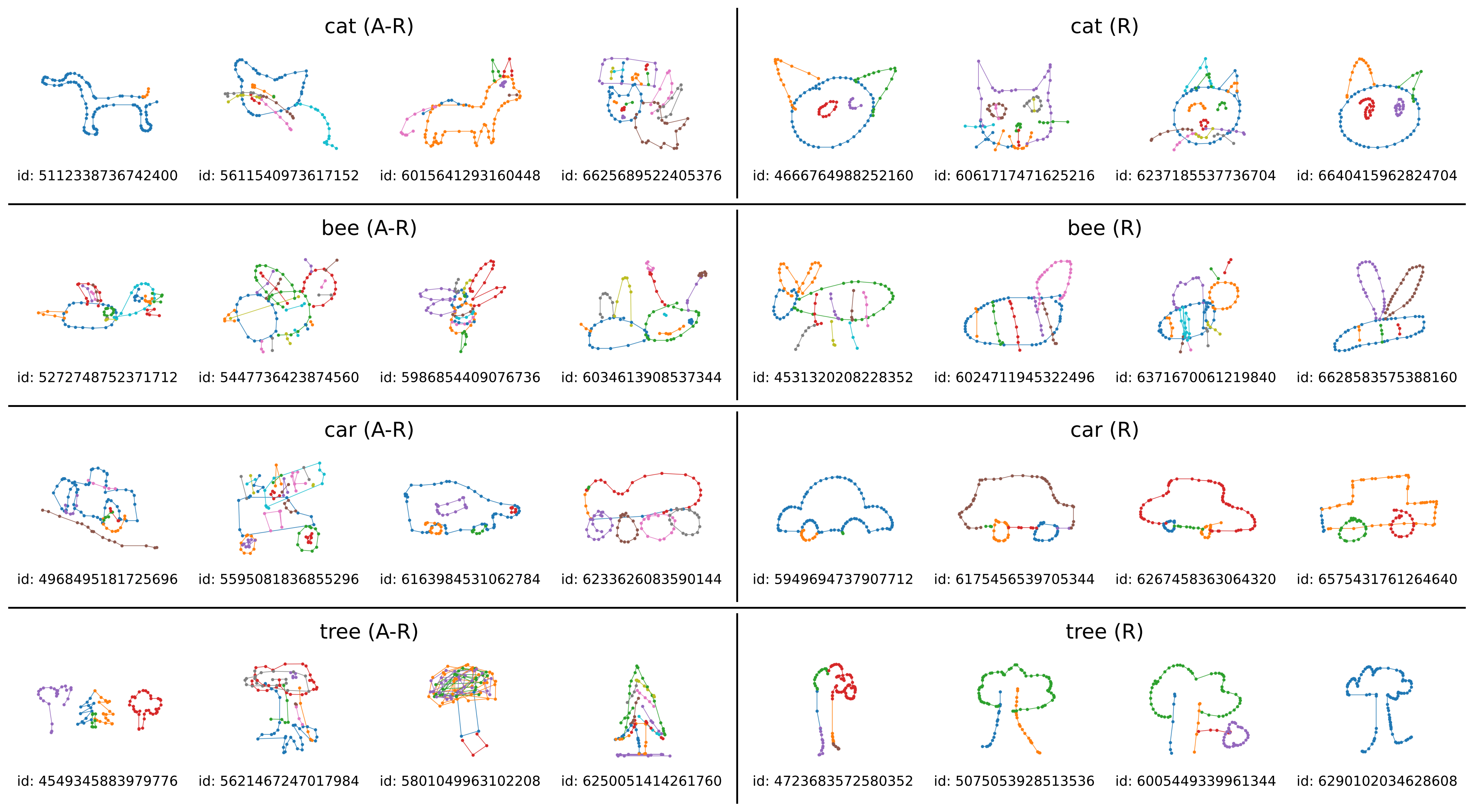}
\caption{Mutually Exclusive Visualization of A$-$R and R Samples in the SketchGraph Dataset. }\label{fig:5}
\end{figure}

Figure~\ref{fig:5} presents a visual comparison between the two dataset variants. The left panel shows samples from the set difference A$-$R, while the right panel shows samples from Version R. For each category, four sketches are randomly selected. Different colors indicate individual strokes, and each sketch is annotated with a unique identifier.

Visually, sketches in Version R show smoother stroke trajectories and more coherent geometric structures, whereas samples from A$-$R contain more fragmented paths and irregular drawing patterns. These visual differences are consistent with the structural statistics reported in Fig.~\ref{fig:3} and Fig.~\ref{fig:4}.

\subsection*{Dataset Properties Summary}

The \textbf{SketchGraph} dataset is characterized by the following properties:

\begin{itemize}
    \item \textbf{Scale and category coverage.} The dataset contains 344 categories and a total of 3.44 million sketch graphs.
    \item \textbf{Spatiotemporal node features.} Each node includes a normalized temporal attribute that encodes stroke order, together with spatial coordinates.
    \item \textbf{Data storage format.} SketchGraph uses binary serialization with offset-based indexing to support batch-level access during training.
\end{itemize}

These properties summarize the structure and composition of the SketchGraph dataset.

\section{Results and Discussion}
\label{sec4}

\subsection{Baseline Configurations and Comparative Framework}
\label{subsec4.1}

To evaluate SketchGraphNet, we compare it with representative baselines from convolutional, sequential, graph-based, and Transformer-style paradigms under a unified experimental configuration.

For raster-based modeling, we include InceptionV3~\cite{szegedy2016rethinking} and MobileNetV2~\cite{sandler2018mobilenetv2}. Sketch graphs are rendered into pixel images using a Bresenham-based rasterization pipeline, following standard input resolutions ($299\times299$ and $224\times224$, respectively).

For sequential modeling, we evaluate BiLSTM~\cite{graves2012long} and BiGRU~\cite{cho2014properties}, which process sketches as ordered stroke sequences.

For graph-native modeling, we compare against commonly used message-passing operators, including GCNConv~\cite{kipf2016semi}, GraphConv~\cite{morris2019weisfeiler}, ChebConv~\cite{defferrard2016convolutional}, SAGEConv~\cite{hamilton2017inductive}, DenseSAGEConv~\cite{fey2019fast}, GINConv~\cite{xu2018powerful}, and GATConv~\cite{velickovic2017graph}. We also include sketch-specific graph architectures such as S3Net~\cite{yang2020s3net}, SketchGNN~\cite{yang2021sketchgnn}, and ESA~\cite{buterez2025end}.

As a Transformer-style graph baseline, we adopt MGT~\cite{xu2021multigraph}, which incorporates global self-attention while remaining trainable under the same batch-size and precision settings. Architectures such as Graphormer~\cite{ying2021transformers} and Exphormer~\cite{shirzad2023exphormer} are closely related; however, their computational overhead prevents controlled comparison under the large-scale sketch setting considered here.

All models are trained and evaluated on SketchGraph using identical data splits, optimization settings, and hardware configurations.

This study focuses on graph-native sketch classification, where sketches are directly modeled as structured graphs. LLMs and vision-language foundation models are not designed to operate on graph-structured sketch inputs; applying them would require rasterization or textual abstraction, which deviates from the formulation considered here. Accordingly, we compare against convolutional, sequential, graph neural, and Transformer-style graph baselines under comparable input representations.

\subsection{Parameter Settings}
\label{subsec4.2}

All experiments are conducted on a single RTX 4070 Ti GPU with 32~GB of system memory. Unless otherwise specified, the batch size is set to 256, the number of training epochs is 7, and the dropout rate is 0. SketchGraphNet uses four ConvBlocks, adopts GINConv as the local message-passing operator, and employs the proposed MemEffAttn as the global attention module with eight attention heads.

Training uses automatic mixed precision~\cite{micikevicius2017mixed}. Adam is used as the optimizer with an initial learning rate of 0.0005 and a cosine annealing schedule. The dataset is split into training, validation, and test sets with a ratio of 9:0.5:0.5.

All baseline models except S3Net are trained using mixed precision. S3Net is trained in FP32 due to sensitivity to low-precision gradients. For each method, results are reported as the best performance over three independent runs; the observed variance is within $\pm 0.15\%$.

\subsection{Main Results and Analysis}
\label{subsec4.3}

We examine the feasibility of applying graph Transformer architectures to corpus-scale sketch recognition. In the case of Exphormer, the construction of expander-graph topologies for a dataset of 3.44 million samples introduces substantial preprocessing overhead, primarily from graph construction and data preparation. Under a single-GPU workstation setting, this overhead prevents a controlled and reproducible end-to-end training procedure. As a result, we restrict our experimental comparison to baseline methods that can be trained under a unified configuration. Table~\ref{tab:1} reports classification accuracy on SketchGraph-A and SketchGraph-R, together with inference latency measured as the average runtime over 100 forward passes with a batch size of 1.

\begin{table*}[t]
\centering
\caption{Classification Accuracy on Both Dataset Versions.}
\label{tab:1}

\resizebox{\textwidth}{!}{%
\begin{tabular}{lcccccccccc}
\toprule
\multirow[c]{2}{*}{Models} &
\multirow[c]{2}{*}{\makecell[c]{Param\\(M)}} &
\multirow{2}{*}{\makecell[c]{Train\\time}} &
\multirow{2}{*}{\makecell[c]{Latency\\(ms)}} &
\multirow{2}{*}{DSSG} &
\multicolumn{3}{c}{SketchGraph-A} &
\multicolumn{3}{c}{SketchGraph-R} \\
\cmidrule(lr){6-8} \cmidrule(lr){9-11}
 & & & & & TOP-1 & TOP-3 & TOP-5 & TOP-1 & TOP-3 & TOP-5 \\
\midrule

\multirow{2}{*}{S3Net}
 & \multirow{2}{*}{5.33} & 1.410 h & 4.206 & w/o & 0.8024 & 0.9246 & 0.9465 & 0.8496 & 0.9640 & 0.9804 \\
 &                       & 1.405 h & 4.237 & w/  & 0.8070 & 0.9272 & 0.9486 & 0.8544 & 0.9663 & 0.9820 \\
\cmidrule(lr{.02em}){1-11}

\multirow{2}{*}{SketchGNN}
 & \multirow{2}{*}{0.79} & 1.047 h & 5.208 & w/o & 0.6936 & 0.8551 & 0.8934 & 0.7457 & 0.9076 & 0.9425 \\
 &                       & 1.058 h & 5.132 & w/  & 0.6986 & 0.8582 & 0.8964 & 0.7511 & 0.9105 & 0.9447 \\
\cmidrule(lr{.02em}){1-11}

\multirow{2}{*}{MGT}
 & \multirow{2}{*}{39.98} & 6.251 h & 6.086 & w/o & 0.7029 & 0.8625 & 0.9009 & 0.7571 & 0.9164 & 0.9495 \\
 &                        & 6.182 h & 6.138 & w/  & 0.6964 & 0.8602 & 0.8990 & 0.7405 & 0.9059 & 0.9431 \\
\cmidrule(lr{.02em}){1-11}

\multirow{2}{*}{ESA}
 & \multirow{2}{*}{6.15} & 2.891 h & 3.739 & w/o & 0.7418 & 0.8960 & 0.9280 & 0.7845 & 0.9317 & 0.9597 \\
 &                       & 1.560 h & 3.772 & w/  & 0.7836 & 0.9168 & 0.9422 & 0.8181 & 0.9512 & 0.9726 \\
\cmidrule(lr{.02em}){1-11}

InceptionV3
 & 27.16 & 13.068 h & 12.614 & w/o & 0.8207 & 0.9367 & 0.9558 & 0.8587 & 0.9687 & 0.9829 \\
\cmidrule(lr{.02em}){1-11}

MobileNetV2
 & 3.50 & 5.040 h & 5.429 & w/o & 0.8113 & 0.9327 & 0.9527 & 0.8542 & 0.9682 & 0.9831 \\
\cmidrule(lr{.02em}){1-11}

BiLSTM
 & 8.78 & 0.771 h & 3.085 & w/o & 0.7924 & 0.9209 & 0.9446 & 0.8465 & 0.9642 & 0.9812 \\
\cmidrule(lr{.02em}){1-11}

BiGRU
 & 6.67 & 0.663 h & 2.968 & w/o & 0.7927 & 0.9205 & 0.9442 & 0.8462 & 0.9647 & 0.9815 \\
\cmidrule(lr{.02em}){1-11}

\multirow{2}{*}{SketchGraphNet}
 & \multirow{2}{*}{8.60} & 1.494 h & 5.610 & w/o & \textbf{0.8313} & \textbf{0.9421} & \textbf{0.9597} & \textbf{0.8731} & \textbf{0.9750} & \textbf{0.9872} \\
 &                       & 1.431 h & 5.642 & w/  & \textbf{0.8362} & \textbf{0.9444} & \textbf{0.9610} & \textbf{0.8761} & \textbf{0.9759} & \textbf{0.9877} \\

\bottomrule
\end{tabular}%
}
\end{table*}

Table~\ref{tab:1} summarizes classification accuracy on SketchGraph-A and SketchGraph-R, together with training time and inference latency. Across both dataset variants, SketchGraphNet achieves the best overall Top-1/Top-3/Top-5 performance, indicating that the proposed local--global design is effective for sparse and structurally heterogeneous sketch graphs. With DSSG enabled, SketchGraphNet reaches 87.61\% Top-1 accuracy on SketchGraph-R and 83.62\% on SketchGraph-A, demonstrating consistent gains under both recognized and noisy sketch distributions.

Beyond accuracy, SketchGraphNet maintains a favorable accuracy--efficiency trade-off. Under the same hardware and batch-size configuration, it trains at roughly 1.4 hours per epoch and runs at about 5.6 ms per sample, while using a substantially smaller parameter budget than the Transformer baseline (MGT). Compared with raster-based CNN baselines, SketchGraphNet avoids image rendering as a preprocessing dependency and attains competitive inference latency under the same evaluation protocol. Together, these results show that SketchGraphNet scales effectively to corpus-level sketch training while retaining strong predictive performance.

\begin{figure}[H]
\centering
\includegraphics[width=\linewidth]{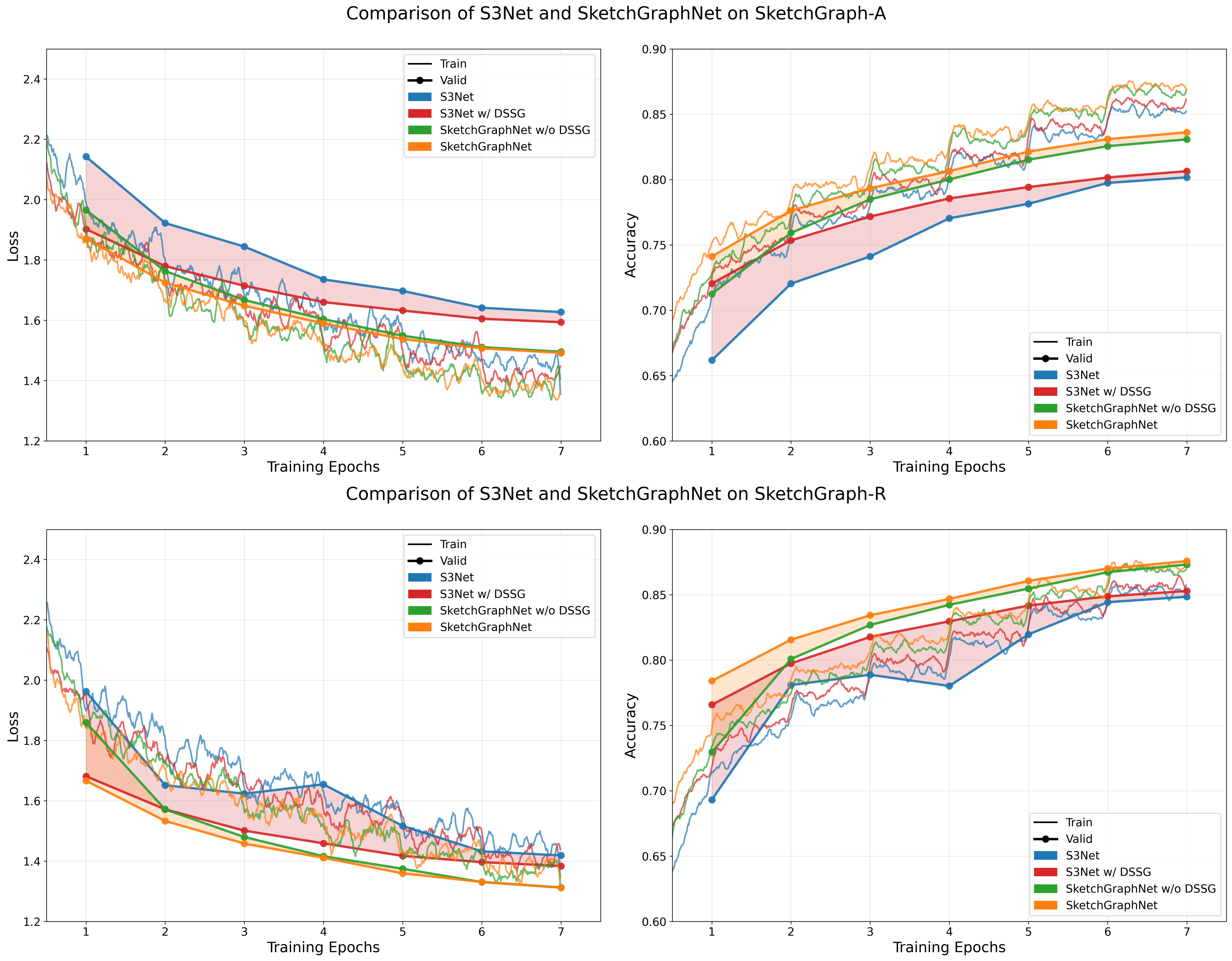}
\caption{Training Accuracy and Loss Curves.}\label{fig:6}
\end{figure}

The results also reflect the influence of DSSG-based structural enhancement. Across multiple models, enabling DSSG is associated with higher classification accuracy. For SketchGraphNet, the absolute accuracy gain from DSSG is relatively modest; however, the training curves in Fig.~\ref{fig:6} show reduced fluctuations compared with the variant without DSSG. This observation suggests that DSSG contributes to more stable training behavior, particularly on the noisier SketchGraph-A variant.

Given that S3Net has a comparable parameter scale and is designed specifically for sketch-graph modeling, we use it as the primary reference for training dynamics visualization. Figure~\ref{fig:6} compares the training accuracy and loss curves of SketchGraphNet and S3Net. Under the same training configuration, SketchGraphNet exhibits a faster reduction in training loss and smoother accuracy curves over epochs.

\begin{figure}[t]
\centering
\includegraphics[
  height=0.6\textheight,
  keepaspectratio]{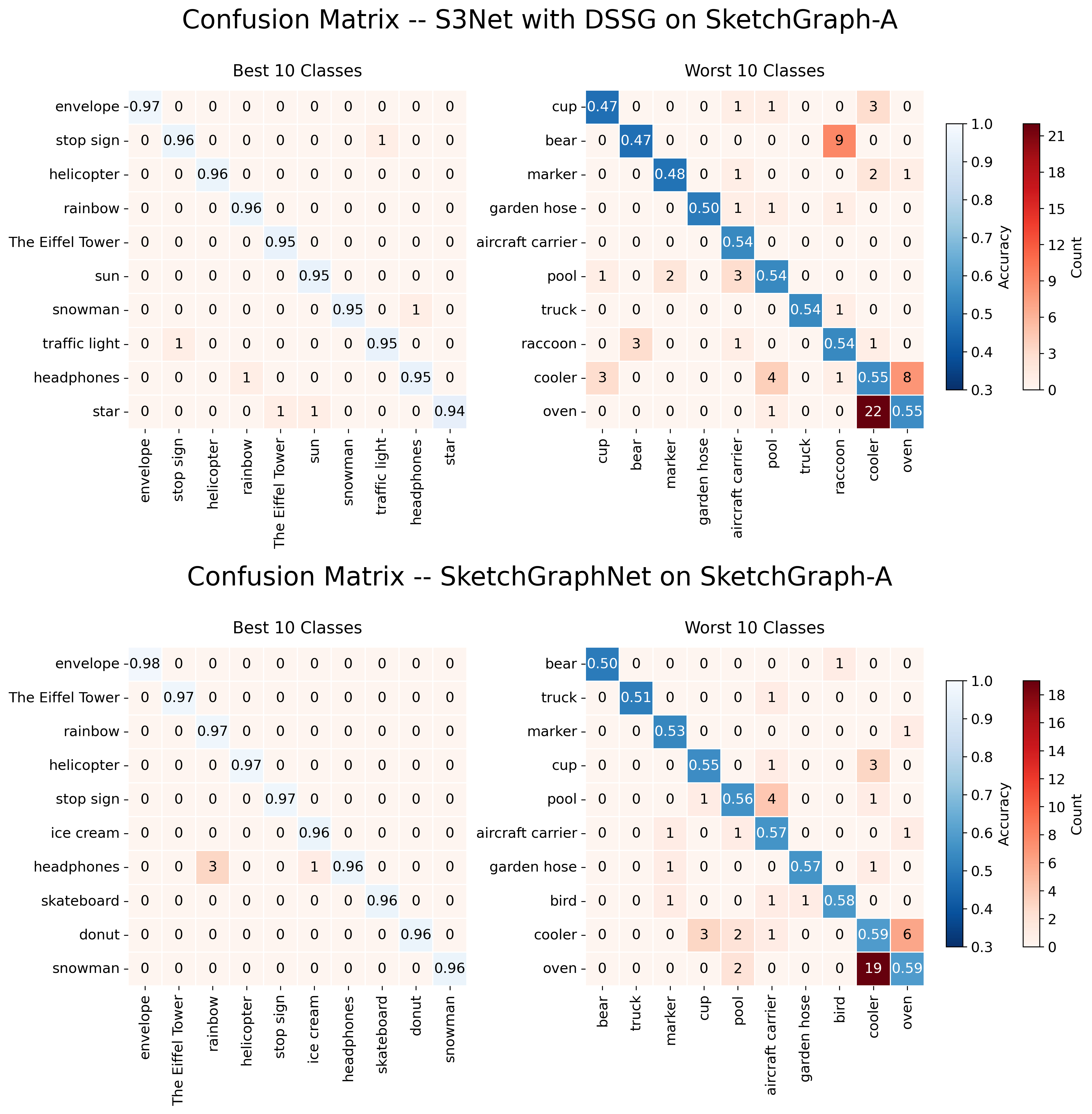}
\caption{Confusion Matrices for Best and Worst Performing Categories.}\label{fig:7}
\end{figure}

On SketchGraph-A, both models exhibit mild overfitting, as indicated by the gap between training and validation performance. Figure~\ref{fig:7} further presents the confusion matrices of SketchGraphNet and S3Net on this dataset.

\FloatBarrier

In Fig.~\ref{fig:7}, higher-intensity colors correspond to stronger inter-class confusion. Categories with close semantic meanings, such as ``cup'' versus ``coffee cup'', exhibit notable confusion, and frequent misclassifications are also observed between categories such as ``cooler'' and ``oven''. These patterns reflect inherent semantic overlap in the dataset taxonomy.

\begin{figure}[t]
\centering
\includegraphics[width=\linewidth]{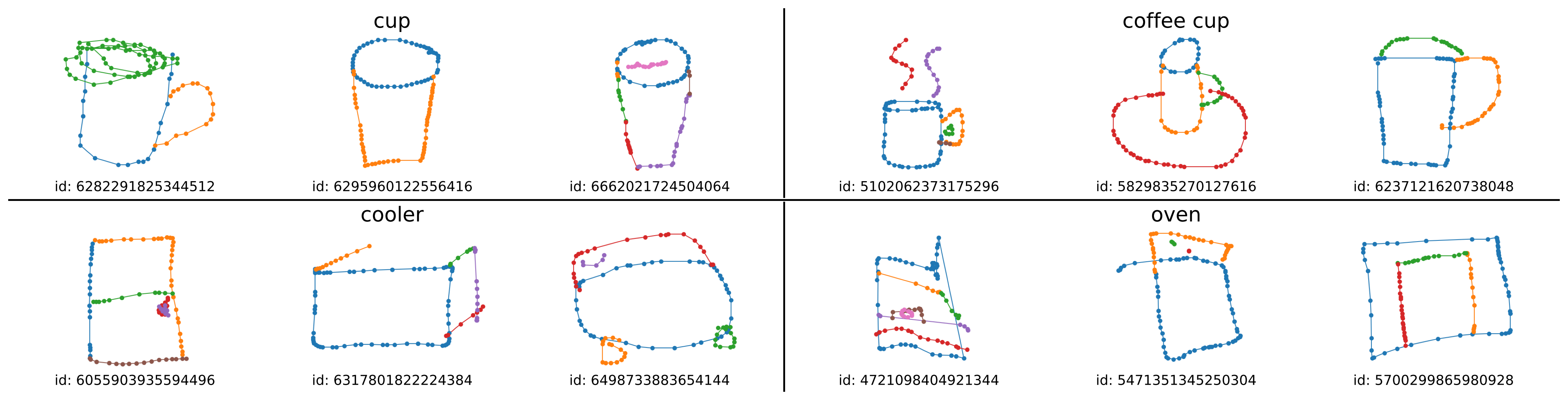}
\caption{Visualization of Easily Confused Categories.}\label{fig:8}
\end{figure}

Comparing the two confusion matrices, SketchGraphNet shows reduced confusion in several closely related categories relative to S3Net with DSSG. Figure~\ref{fig:8} presents qualitative examples of such categories to further illustrate typical failure cases.

\subsection{Ablation Studies and Efficiency Analysis}
\label{subsec4.4}

\subsubsection{Efficiency vs. Architectural Depth}
\label{subsec4.4.1}

We conduct ablation studies to examine the effects of architectural depth and global attention design on classification performance, training efficiency, and numerical stability. In these experiments, MemEffAttn refers to a Softmax-based self-attention module in which a non-negative ReLU mapping is applied to the query and key projections before attention computation. The attention weights are computed using exact Softmax and are executed with a tiled, blockwise strategy based on xFormers to control peak memory usage.

Table~\ref{tab:2} examines the impact of varying the number of ConvBlocks and compares the proposed MemEffAttn with PerformerAttention~\cite{choromanski2020rethinking}. Both attention mechanisms are designed to address memory constraints in self-attention, but they differ in implementation. PerformerAttention employs random-feature approximations to avoid explicit attention matrix construction, whereas MemEffAttn computes exact Softmax attention and adopts a tiled, blockwise execution strategy with ReLU-mapped query and key projections.

\begin{table}[t]
\centering
\caption{Evaluation of ConvBlock Count and Global Attention Methods.}
\label{tab:2}

\resizebox{\columnwidth}{!}{%
\begin{tabular}{lccccc}
\toprule
\multirow{2}{*}{\makecell[l]{Global\\Attention}} &
\multirow{2}{*}{\makecell[c]{Conv\\Blocks}} &
\multirow{2}{*}{\makecell[c]{Avg.\\Memory}} &
\multirow{2}{*}{\makecell[c]{Avg.\\Train Time}} &
\multicolumn{2}{c}{Top1 Accuracy} \\
\cmidrule(lr){5-6}
 & & & & SketchGraph-A & SketchGraph-R \\
\midrule

\multirow{3}{*}{PerformerAttention}
& 4 & 5.03 GB & 2.12 h & 0.8344 & 0.8756 \\
& 6 & 6.44 GB & 3.03 h & 0.8404 & 0.8804 \\
& 8 & 7.93 GB & 3.93 h & 0.8423 & 0.8823 \\

\cmidrule(lr{.02em}){1-6}

\multirow{3}{*}{MemEffAttn}
& 4 & 2.87 GB & 1.38 h & 0.8362 & 0.8761 \\
& 6 & 3.60 GB & 1.93 h & 0.8414 & 0.8813 \\
& 8 & 4.36 GB & 2.48 h & 0.8432 & 0.8822 \\

\bottomrule
\end{tabular}%
}
\end{table}

As shown in Table~\ref{tab:2}, MemEffAttn is associated with lower average memory usage and shorter training time across all evaluated depths ($L\in\{4,6,8\}$) compared with PerformerAttention. At the same time, the Top-1 accuracy achieved by MemEffAttn remains comparable to that of PerformerAttention on both SketchGraph-A and SketchGraph-R, with small differences observed across different depths.

These results show that varying the number of ConvBlocks affects accuracy, memory consumption, and training time in a consistent manner for both attention mechanisms. Under the same depth settings, MemEffAttn exhibits a more compact memory footprint and reduced training time while maintaining similar classification performance.

\begin{table}[H]
\centering
\small
\caption{Evaluation of Local GNN Operator Replaceability.}
\label{tab:3}
\begin{tabular}{lcccc}
\toprule
\multirow{2}{*}{Local GNN} &
\multirow{2}{*}{DSSG} &
\multicolumn{2}{c}{TOP-1 Accuracy} \\
\cmidrule(lr){3-4}
& & SketchGraph-A & SketchGraph-R \\
\midrule

\multirow{2}{*}{GINConv}
& w/o & 0.8313 & 0.8731 \\
& w/  & 0.8362 & 0.8761 \\
\cmidrule(lr{.02em}){1-4}

\multirow{2}{*}{DenseSAGEConv}
& w/o & 0.8312 & 0.8709 \\
& w/  & 0.8337 & 0.8753 \\
\cmidrule(lr{.02em}){1-4}

\multirow{2}{*}{SAGEConv}
& w/o & 0.8302 & 0.8725 \\
& w/  & 0.8335 & 0.8757 \\
\cmidrule(lr{.02em}){1-4}

\multirow{2}{*}{GraphConv}
& w/o & 0.8317 & 0.8714 \\
& w/  & 0.8343 & 0.8757 \\
\cmidrule(lr{.02em}){1-4}

\multirow{2}{*}{GCNConv}
& w/o & 0.8312 & 0.8718 \\
& w/  & 0.8342 & 0.8748 \\
\cmidrule(lr{.02em}){1-4}

\multirow{2}{*}{GATConv}
& w/o & 0.8342 & 0.8734 \\
& w/  & 0.8337 & 0.8740 \\
\cmidrule(lr{.02em}){1-4}

\multirow{2}{*}{ChebConv}
& w/o & 0.8315 & 0.8718 \\
& w/  & 0.8249 & 0.8678 \\

\bottomrule
\end{tabular}
\end{table}

\subsubsection{Local and Global Component Ablation}
\label{subsec4.4.2}
To examine the effect of different local message-passing operators, we replace the local GNN backbone in SketchGraphNet with several commonly used GNN variants. Table~\ref{tab:3} reports the Top-1 accuracy obtained with each operator, both with and without DSSG.

Across different local GNN choices, the classification accuracy varies within a relatively narrow range on both SketchGraph-A and SketchGraph-R. For most operators, enabling DSSG is associated with higher Top-1 accuracy, although the magnitude of improvement differs across architectures. These results indicate that the overall performance trends are not tied to a specific local GNN choice.

We further isolate the contributions of the global attention branch and the temporal feature $t$. As shown in Table~\ref{tab:4}, removing the temporal feature is associated with a consistent decrease in Top-1 accuracy on both dataset variants. Removing the global attention branch leads to a larger reduction in classification accuracy across the two datasets.

\begin{table}[!t]
\centering
\small
\caption{Ablation Study on Global Attention Structures.}
\label{tab:4}

\begin{tabular}{lccc}
\toprule
\multirow{2}{*}{Configuration} & \multicolumn{2}{c}{TOP-1 Accuracy} \\
\cmidrule(lr){2-3}
& SketchGraph-A & SketchGraph-R \\
\midrule

Full SketchGraphNet & \textbf{0.8362} & \textbf{0.8761} \\
\cmidrule(lr{.02em}){1-3}

w/o Global Attention & 0.7589 & 0.8048 \\
\cmidrule(lr{.02em}){1-3}

w/o Temporal Feature & 0.8171 & 0.8632 \\

\bottomrule
\end{tabular}
\end{table}

Table~\ref{tab:4} reports the ablation results obtained by removing the global attention branch and the temporal feature $t$, respectively. When the global attention branch is removed, the classification accuracy decreases substantially on both SketchGraph-A and SketchGraph-R. Removing the temporal feature also leads to a reduction in Top-1 accuracy, although the magnitude of the decrease is smaller than that observed for removing global attention.

These results show that both components contribute to overall classification performance under the same training configuration, with different levels of impact reflected in the observed accuracy changes.

\subsubsection{Numerical Stability and Attention Design}
\label{subsec4.4.3}

We further examine the effect of applying a non-negative ReLU mapping to the query and key projections. Table~\ref{tab:5} reports the results obtained with and without this mapping under different network depths.

\begin{table}[!t]
\centering
\small
\caption{Ablation Study on Nonlinear-Kernels.}
\label{tab:5}
\setlength{\tabcolsep}{12pt}

\begin{tabular}{lcccc}
\toprule
\multirow{2}{*}{MemEffAttn} &
\multirow{2}{*}{\makecell[c]{Number of\\ConvBlock}} &
\multicolumn{2}{c}{TOP-1 Accuracy} \\
\cmidrule(lr){3-4}
 & & SketchGraph-A & SketchGraph-R \\
\midrule

\multirow{3}{*}{w/ Kernel}
& 4 & 0.8362 & 0.8761 \\
& 6 & 0.8419 & 0.8815 \\
& 8 & 0.8432 & 0.8822 \\
\cmidrule(lr{.02em}){1-4}

\multirow{3}{*}{w/o Kernel}
& 4 & 0.8376 & 0.8773 \\
& 6 & 0.8430 & 0.8584 \\
& 8 & 0.0031 & 0.0028 \\

\bottomrule
\end{tabular}
\end{table}

When the ReLU mapping is removed, the model exhibits increasing instability as the number of ConvBlocks grows. In particular, at eight layers, training diverges under mixed-precision settings, and Inf values are observed in the $\boldsymbol{Q}$ and $\boldsymbol{K}$ tensors. In contrast, configurations with the ReLU mapping remain trainable across all evaluated depths and achieve stable accuracy on both dataset variants.

Finally, Table~\ref{tab:6} reports a controlled stress test comparing MemEffAttn with a standard Softmax attention implementation under the same ReLU-based query and key mapping. In both configurations, attention weights are computed using exact Softmax, and the two implementations differ only in their execution strategy.

\begin{table}[h]
\centering
\small
\caption{Stability analysis under identical ReLU-kernel mapping.}
\label{tab:6}
\setlength{\tabcolsep}{8pt}
\renewcommand{\arraystretch}{1.3}

\begin{tabular}{lcccc}
\toprule
\multirow{2}{*}{\makecell[l]{Attention\\Implementation}} &
\multirow{2}{*}{\makecell[c]{Implementation\\Type}} &
\multirow{2}{*}{\makecell[c]{Peak\\Memory}} &
\multirow{2}{*}{\makecell[c]{Convergence\\Status}} &
\multirow{2}{*}{\makecell[c]{First NaN\\Epoch}} \\
& & & & \\
\midrule

Standard Attention & \makecell[c]{Explicit NN\\Matrix} & 4.07 GB & Failed & 1 \\
\cmidrule(lr{.02em}){1-5}

MemEffAttn & \makecell[c]{Tiling-based\\(xFormers)} & 2.87 GB & Stable & None \\

\bottomrule
\end{tabular}
\end{table}

Under automatic mixed-precision training, the standard attention implementation becomes numerically unstable and produces NaN loss during the first training epoch. In contrast, MemEffAttn remains stable throughout training and exhibits a lower peak memory footprint under the same configuration.

\subsubsection{Hyperparameter Sensitivity}
\label{subsec4.4.4}

We further evaluate the sensitivity of SketchGraphNet to several key hyperparameters, including learning rate, label smoothing, and dropout. Table~\ref{tab:7} reports the Top-1 accuracy obtained under different parameter settings.

\begin{table}[!t]
\centering
\small
\caption{Parameter Sensitivity Analysis of SketchGraphNet.}
\label{tab:7}
\setlength{\tabcolsep}{10pt}
\renewcommand{\arraystretch}{1.2}

\begin{tabular}{lccc}
\toprule
\multirow{2}{*}{SketchGraphNet} & \multicolumn{2}{c}{TOP-1 Accuracy} \\
\cmidrule(lr){2-3}
& SketchGraph-A & SketchGraph-R \\
\midrule

Default settings & 0.8362 & 0.8761 \\
\cmidrule(lr{.02em}){1-3}

Learning Rate (0.0001) & 0.8136 & 0.8575 \\
\cmidrule(lr{.02em}){1-3}

Label Smoothing (0.0) & 0.8317 & 0.8741 \\
\cmidrule(lr{.02em}){1-3}

DropOut (0.2) & 0.8291 & 0.8698 \\

\bottomrule
\end{tabular}
\end{table}

Reducing the learning rate results in lower classification accuracy on both dataset variants. Disabling label smoothing also leads to a decrease in accuracy compared with the default configuration. Introducing dropout with a rate of 0.2 yields a modest performance reduction under the same training setup. Compared with optimization-related parameters such as learning rate and label smoothing, the observed accuracy variations under different regularization settings remain relatively limited.

\subsection{Theoretical and Practical Implications}
\label{subsec4.implications}

From a theoretical perspective, this work contributes to structured information modeling by formalizing free-hand sketches as graph-native objects and demonstrating that local--global hybrid graph architectures can scale to corpus-level training without auxiliary positional encoding. The proposed MemEffAttn illustrates that feature-space stabilization combined with implementation-level memory optimization can improve numerical robustness in hybrid graph Transformer settings.

From a practical perspective, SketchGraphNet enables efficient large-scale training on commodity single-GPU hardware, with reduced memory footprint and stable mixed-precision behavior. The SketchGraph benchmark further provides a reproducible evaluation platform for future graph-based sketch research within the computing and information science community.

\FloatBarrier

\section{Conclusion}
\label{sec5}

This work presents SketchGraphNet, a hybrid graph neural architecture that integrates local message passing with a memory-efficient global attention mechanism for large-scale sketch-graph classification. Evaluated on the proposed SketchGraph benchmark, including both SketchGraph-A and SketchGraph-R, SketchGraphNet achieves consistently higher classification accuracy than representative convolutional, sequential, and graph-based baselines under a unified training configuration. The model attains these results with a compact parameter budget and efficient training and inference characteristics, while remaining stable under mixed-precision training at corpus scale. Together, these findings demonstrate that combining local--global modeling with an implementation-level efficient attention design provides a practical and scalable solution for structured sketch understanding in sparse and noisy real-world settings.

\section*{CRediT authorship contribution statement}

\textbf{Shilong Chen}: Methodology, Theoretical Analysis, Algorithm Design, Writing -- original draft, Writing -- review \& editing. 
\textbf{Mingyuan Li}: Methodology, Theoretical Analysis, Algorithm Design, Writing -- review \& editing. 
\textbf{Zhaoyang Wang}: Supervision, Writing -- review \& editing. 
\textbf{Zhonglin Ye}: Funding acquisition, Writing -- review \& editing. 
\textbf{Haixing Zhao}: Supervision, Writing -- review \& editing.

\section*{Declaration of interests}

The authors declare that they have no known competing financial interests or personal relationships that could have appeared to influence the work reported in this paper.

\section*{Acknowledgments}

This work is supported by Scientific Research Innovation Capability Support Project for Young Faculty (SRICSPYF-BS2025007), Natural Science Foundation of Qinghai Province (2025-ZJ-994M), National Natural Science Foundation of China (62566050).


\bibliographystyle{elsarticle-num} 

\bibliography{refs}

\end{document}